\pdfoutput=1

\documentclass[11pt]{article}

\usepackage[final]{acl}

\usepackage{times}
\usepackage{latexsym}

\usepackage[T1]{fontenc}

\usepackage[utf8]{inputenc}

\usepackage{microtype}

\usepackage{inconsolata}

\usepackage{graphicx}
\usepackage{comment}
\usepackage{paralist}
\usepackage{enumitem} 
\usepackage{amsmath}  
\usepackage{amssymb}   
\usepackage{hyperref} 
\usepackage{soul} 
\usepackage{multirow} 
\usepackage{comment} 
\usepackage{tabularx}
\usepackage{booktabs}
\usepackage{tikz} 
\usepackage{paralist} 
\usepackage{adjustbox}
\usepackage{verbatim}
\usepackage{xcolor} 
\usepackage{bigstrut}
\definecolor{darkgreen}{rgb}{0.0, 0.5, 0.0}
\usepackage{authblk}

\newcommand{\bi}{\begin{itemize}}
\newcommand{\ei}{\end{itemize}}
\newcommand{\BE}{\begin{enumerate}}
\newcommand{\EE}{\end{enumerate}}

\newcommand{\etc}{\mbox{\it etc.}}

\newcommand{\ie}{\mbox{\it i.e.}}

\newcommand{\Ie}{\mbox{\it I.e.}}
\newcommand{\wrt}{\mbox{\it w.r.t.}}

\newcommand{\viz}{\mbox{\it viz.}}

\newcommand{\ourmethodshort}{\textsc{PA-RAG}}

\newcommand{\ourmethodlong}{Paraphrase Augmentation for Retrieval-Augmented Generation}

\newcommand{\maxtokens}{T}
\newcommand{\numqacalls}{N_c}

\newcommand{\corruptionprob}{\mathrm{p}}

\newcommand{\raftmix}{\text{\sc CA-RAFT}}

\newcommand{\raftmixfn}[1]{\textsc{CA-RAFT}(#1)}

\newcommand{\chaptercorruption}{Ch.w. bucket}
\newcommand{\querywisecorruption}{Q.w. bucket }

\title{Systematic Knowledge Injection into Large Language Models via Diverse Augmentation for Domain-Specific RAG
}

\author[1]{Kushagra Bhushan\thanks{Equal contribution.}\thanks{This work was conducted while Kushagra Bhushan was an intern at IBM Research.}}
\author[2]{Yatin Nandwani\textsuperscript{*}}
\author[2]{Dinesh Khandelwal}
\author[2]{Sonam Gupta\textsuperscript{2},\authorcr Gaurav Pandey}
\author[2]{Dinesh Raghu}
\author[2]{Sachindra Joshi}

\affil[ ]{\begin{minipage}{\textwidth}
\centering 
\texttt{20je0506@mc.iitism.ac.in, \{yatin.nandwani@, dikhand1@in., sonam.gupta7@, } \newline
\texttt{gpandey1@in.,diraghu1@in., jsachind@in.\}ibm.com} \newline \newline
\textsuperscript{1}IIT(ISM) Dhanbad \quad \textsuperscript{2}IBM Research
\end{minipage}
}

\begin{document}
\maketitle
\begin{abstract}
Retrieval-Augmented Generation (RAG) has emerged as a prominent method for incorporating domain knowledge into Large Language Models (LLMs).
While RAG enhances response relevance by incorporating retrieved domain knowledge in the context, retrieval errors can still lead to hallucinations and incorrect answers. 
To recover from retriever failures, domain knowledge is injected by fine-tuning the model to generate the correct response, even in the case of retrieval errors. 
However, we observe that without systematic knowledge augmentation, fine-tuned LLMs may memorize new information but still fail to extract relevant domain knowledge, leading to poor performance.
In this work, we present a novel framework that significantly enhances the fine-tuning process by augmenting the training data in two ways -- context augmentation and knowledge paraphrasing. 
In context augmentation, we create multiple training samples for a given QA pair by varying the relevance of the retrieved information, teaching the model when to ignore and when to rely on retrieved content.
In knowledge paraphrasing, we fine-tune with multiple answers to the same question, enabling LLMs to better internalize specialized knowledge. 
To mitigate catastrophic forgetting due to fine-tuning,
we add a domain-specific identifier to a question and also utilize a replay buffer containing general QA pairs. 
Experimental results demonstrate the efficacy of our method over existing techniques, achieving up to 10\% relative gain in token-level recall while preserving the LLM's generalization capabilities.

\end{abstract}
\section{Introduction} \label{sec:intro}
Large Language Models (LLMs) have transformed natural language processing, excelling across various tasks~\citep{NEURIPS2020_1457c0d6}. To tailor LLMs for domain-specific applications, such as question answering over specialized corpora, Retrieval-Augmented Generation (RAG) has emerged as a popular approach~\citep{lewis2020retrieval, karpukhin2020dense}. Although RAG enhances the relevance of generated answers, it is prone to hallucinations \citep{ji-etal-2023-survey, nandwani-etal-2023-pointwise}, especially when the retriever fails to fetch relevant documents from the corpus.

To address this issue, \textit{knowledge injection} techniques~\citep{zhang2024raft,yoran2023making} have been proposed. 
Knowledge injection has two main objectives: (1) when the retriever succeeds in fetching correct documents, then the LLM should be able to leverage that information to generate an appropriate response, and (2) when the retriever fails, the LLM should recall the domain specific information from the infused parametric knowledge to generate the response.
To achieve them, existing techniques fine-tune LLMs with domain-specific data, embedding the knowledge directly into the LLM's parameters. Specifically, these techniques fine-tune LLMs to generate the correct response to a question irrespective of the relevance of the retrieved documents.
However, existing knowledge injection techniques suffer from two main issues:
\begin{compactenum}[1)]
   \item \textit{Conditional Memorization Bias:} In the training data, each question is assigned to either a retrieval success or a retrieval failure scenario. 
   \Ie, the relevant information to answer the question is either present or absent in the given context for a question.
   This static assignment determines how the LLM learns the knowledge.
   For example, consider a scenario where all training questions from a particular document are assigned to the `retrieval success' bucket, \ie, information required to answer is present in the accompanying context.
   In this case, 
   the LLM is encouraged to rely on the external retrieved context, and it may not memorize information from such documents.
   Conversely, if all training questions from a particular document are assigned to `retrieval failure' bucket, the LLM will be forced to memorize its content during fine-tuning and may learn to ignore the provided information for questions from that document.
   As a result, the LLM learns different sections of the domain data in different ways. This inconsistency can be problematic, as the LLM might struggle when faced with opposite scenarios during inference. We confirm this behaviour in our experiments.

	\item \textit{Canonical Answer Overfitting:} Each question in the fine-tuning dataset is associated with only one canonical answer. This singular association leads the LLM to learn and replicate spurious patterns~\cite{allen2024physics}, treating the answer as a fixed representation for that specific question. As a result, the LLM's ability to generate nuanced or diverse responses based on varying contextual factors is constrained.
\end{compactenum}

To mitigate these issues, we propose \ourmethodshort: \ourmethodlong, a novel fine-tuning framework that improves knowledge injection into LLMs for domain-specific RAG tasks. 
\ourmethodshort\ introduces two different ways of training data augmentation to mitigate conditional memorization bias and canonical answer overfitting.

First,
\ourmethodshort\
uses context augmentation to simulate both retrieval success and retrieval failure scenarios for all the training questions. 
This prevents conditional memorization bias by teaching the model to identify whether the given retrieved information is relevant or not and then accordingly decide to ignore or utilize it.

Second, \ourmethodshort\ synthetically generates multiple answers for each training question to mitigate canonical answer overfitting.
It is inspired by a recent study~\citep{allen2024physics} demonstrating that paraphrasing knowledge during pre-training significantly enhances LLMs' ability to inject and extract that information in the downstream tasks. 
However, they focus on paraphrasing only during pre-training and not during fine-tuning.
Building on this, we propose to systematically augment the fine-tuning data by synthetically generating multiple answers for each question. This encourages LLMs to memorize and extract the domain knowledge effectively while minimizing the overfitting on stylistic features.

Furthermore, to address the challenge of \emph{catastrophic forgetting}~\citep{zhang2024dissecting, ke2023continual, jang2021towards}—where an LLM’s general language understanding deteriorates as domain-specific fine-tuning overwrites prior knowledge—we introduce a \emph{self-selective rehearsal replay buffer}~\citep{gupta2024selective, huang2024mitigating}.
This buffer contains samples from an instruction tuning dataset. But unlike traditional replay buffers, the self-selective approach uses LLM’s own predictions rather than the gold response to retain the LLM’s versatility across tasks when injecting knowledge.

In addition, we also introduce the novel use of \emph{Domain Identifiers}—phrases pre-pended to questions—to help the LLM distinguish
new knowledge from its existing skills.
This improves domain-specific accuracy and also mitigates catastrophic forgetting, preserving LLM's general competence during fine-tuning. 

To validate \ourmethodshort's ability to inject new domain-specific knowledge, we require a corpus that the LLM hasn't seen during pre-training or instruction tuning. 
In the absence of such a corpus, we create two datasets using domain-specific books that were published in 2024\footnote{after the cutoff date of the fine-tuned LLM}. 
In our experiments, \ourmethodshort\ achieves a significant increase in performance against other methods on this domain-specific benchmark while maintaining general reasoning capabilities that we measure using various benchmarks such as MMLU \cite{hendrycks2020measuring}, TruthfulQA \cite{lin2022truthfulqa}, Hellaswag \cite{zellers2019hellaswag}, and GSM8k \cite{cobbe2021gsm8k}.

\section{Related Work}\label{sec:related}

\noindent\textbf{Retrieval Augmented Generation:} RAG enhances Large Language Models (LLMs) by integrating external data sources, such as knowledge bases, to improve relevance and accuracy~\citep{lewis2020retrieval, guu2020retrieval, karpukhin2020dense}. Recent advancements have extended its applicability across domains~\citep{asai2024self, kimsure, yan2024corrective, liu-etal-2024-ra}, but RAG systems still face key challenges: hallucinations due to mismatches between retrieved data and the LLM’s pre-existing knowledge~\citep{setty2024improvingretrievalragbased, jin2024tug}, difficulty with complex multi-document reasoning~\citep{setty2024improvingretrievalragbased}, and an inability to fully leverage fixed-domain settings where all domain-specific documents are available beforehand because typically neither the retriever nor the generator LLM are trained on the domain data.

\noindent\textbf{Domain-Aware Fine-Tuning for RAG:} Joint training of the retriever and LLM has been proposed as a way to improve RAG’s domain-specific performance~\citep{guu2020retrieval, singh2021end, siriwardhana-etal-2023-improving, shi2024replug}. By jointly training the retriever and LLM, the system can better adapt to domain-specific contexts. However, this approach introduces complexities, including the need for specialized loss functions and frequent retriever updates. 

Another line of work~\citep{mecklenburg2024injecting, zhang2024self} focuses solely on adding domain knowledge to LLMs as an alternative to RAG.
These approaches fine-tune LLMs using question-answer (QA) pairs derived from domain data and aim to answer any new test query without retrieving any document. As a result, they fail to leverage access to the domain documents during inference.

Recently,~\citeauthor{zhang2024raft} introduced~\textit{Retrieval-Augmented Fine-Tuning (RAFT)}, a fine-tuning method for LLMs to incorporate domain knowledge and enhance in-domain RAG performance. 
RAFT combines RAG and fine-tuning by training LLMs on domain data using a mixture of oracle and distractor document contexts.
However, it suffers from conditional memorization bias and canonical answer overfitting. 
On the other hand, \ourmethodshort\ uses context augmentation and answer paraphrasing to address these issues.

\noindent\textbf{Catastrophic forgetting:}
Catastrophic Forgetting~\citep{french1999catastrophic, zheng-etal-2024-learn} occurs when new domain-specific fine-tuning overwrites previously learned general knowledge, reducing performance on earlier tasks. Replay-based methods~\citep{de2019episodic,rolnick2019experience}, help mitigate this by rehearsing prior task data during training. Recent advances in replay-based approaches for language models~\citep{scialom2022fine, mok2023large} have shown promise in reducing catastrophic forgetting. The Self-Synthesized Rehearsal (SSR)~\cite{gupta2024selective, huang2024mitigating} framework uses the LLM to generate synthetic rehearsal data, reducing reliance on stored instances.
\section{Problem Definition}
\label{sec:prob_def}

Given a domain-specific corpus $D = \{d_i\}_{i=1}^{n}$, where each document contains domain-relevant knowledge, the goal is to fine-tune Large Language Models (LLMs) for enhanced performance in domain-specific Retrieval-Augmented Generation (RAG) systems. Unlike previous works~\citep{linra, wanginstructretro, mecklenburg2024injecting}, which handle changing test-time domains or documents, our approach, like RAFT, assumes a fixed target domain with known access to domain-specific documents.

Our proposed method, \ourmethodshort\ described in Section~\ref{sec:approach}, addresses the following key challenges:

\begin{compactenum}[1)]

    \item Canonical answer overfitting: In Section~\ref{subsec:synthetic_qa_generation}, we discuss how paraphrased answer augmentation is used to generate multiple variations of answers per train question, ensuring the LLM learns the underlying knowledge and avoids overfitting to fixed answers.
    
	\item Conditional memorization bias: In Section~\ref{subsec:fine_tuning_strategy}, we introduce a method that simulates both successful and failed retrieval scenarios for each question. This helps the LLM learn how to handle various retrieval conditions, preventing it from memorizing answers in static settings and improving its generalization.

	\item Catastrophic forgetting: In Sections~\ref{subsec:domain_indentifiers} and ~\ref{subsec:self_selective_replay}, we outline how a self-selective rehearsal replay buffer and domain-specific identifiers help balance domain-specific knowledge acquisition while retaining the LLM's general knowledge and capabilities.
\end{compactenum}

Through these strategies, \ourmethodshort\ enables more effective knowledge integration, improving domain-specific performance while maintaining generalization capabilities in RAG systems.
\section{Approach}
\label{sec:approach}

In this section, we outline our approach, which involves --
1). generating a synthetic QA dataset from the documents of the given domain;
2). augmenting it to improve knowledge extraction and;
3). incorporating domain identifiers to clearly define boundaries between distinct knowledge areas.

\subsection{Synthetic QA Generation}
\label{subsec:synthetic_qa_generation}
Given a domain-specific corpus $D= \{d_i\}_{i=1}^{n}$ of $n$ documents, our objective is to generate diverse question-answer (QA) pairs that maximize information coverage across the documents using an LLM. We generate these QA pairs using Mixtral-8x22B-Instruct-v0.1\footnote{\href{https://huggingface.co/mistralai/Mixtral-8x22B-Instruct-v0.1}{mistralai/Mixtral-8x22B-Instruct-v0.1}} as the LLM.

\noindent\textbf{Document Chunking:} 
If a document $d_i$ has more than $\maxtokens$ tokens, we split it into chunks of size $\maxtokens//2$. Otherwise, it is processed as a single chunk.
Here $\maxtokens$ is a hyperparameter.
We denote the set of chunks for document $d_i$ as $C^{(i)} = \{c^{(i)}_1, c^{(i)}_2, \dots, c^{(i)}_{m_i}\}$, where $m_i$ is the total number of chunks for document $d_i$. Each $c^{(i)}_s$ represents the $s$-th chunk of document $d_i$, where $1 \leq s \leq m_i$.

\noindent\textbf{QA Pair Generation:} 
For each chunk $c^{(i)}_s \in C^{(i)}$, we prompt the LLM $\numqacalls$ times using an instruction $P_1$ along with the chunk $c^{(i)}_s$. 
Each call to the LLM using prompt $P_1$ generates multiple QA pairs, ensuring that most of the tokens in $c^{(i)}_s$ are covered. Multiple calls with nucleus sampling~\cite{holtzmancurious} ensure adequate token coverage. 
Let $\{(q_s^j, a_s^j)\}_{j=1}^{k}$ be the set of QA pairs generated for chunk $c^{(i)}_s$.
We call this dataset of QA pairs as $\mathcal{D}_{\text{base}} = 
\{ \{(q_s^j, a_s^j)\}_{j=1}^{k}, \forall c^{(i)}_s \in C^{(i)}, \forall d_i \in D\}.$
This approach ensures broad content coverage, but the lack of variability in the generated answers can lead to Canonical Answer Overfitting, as described in Section~\ref{sec:intro}.

\noindent\textbf{Adding Answer Multiplicity:} 
To address canonical answer overfitting, we introduce answer multiplicity by re-prompting the LLM to generate multiple answers for each generated question $q_s^j$ using a separate prompt $P_2$. For each chunk-question pair $(c^{(i)}_s, q_s^j)$, we generate a set of paraphrased answers $\{a_s^{j1}, a_s^{j2}, \dots, a_s^{jp}\}$, where $p$ is the number of paraphrases:\\
$
\mathrm{LLM}(P_2, c^{(i)}_s, q_s^j) \rightarrow \{a_s^{j1}, a_s^{j2}, \dots, a_s^{jp}\}
$ \\
This results in the augmented dataset $\mathcal{D}_{\text{aug.}}$, where each question $q$ is linked to diverse paraphrased answers. The added variability mitigates canonical answer overfitting and improves generalization. As shown in ablation studies, the model fine-tuned on $\mathcal{D}_{\text{aug.}}$ outperforms the one fine-tuned on the base dataset $\mathcal{D}_{\text{base}}$.
All prompts used for dataset creation, including $P_1$ and $P_2$, are provided in Section~\ref{subsec: appendix_prompts}.

\subsection{Fine-tuning Strategy} \label{subsec:fine_tuning_strategy}

In the original RAFT approach~\citep{zhang2024raft}, each data point used for fine-tuning consists of a question ($q$), a context (a collection of documents), and an answer ($a$). The context is classified as either \emph{relevant}, \ie, it contains at least one document (alongside distractor documents) that provides the information needed to deduce the answer, or \emph{irrelevant} where the entire context consists of distractor documents.
Baseline RAFT, referred to as RAFT($\mathrm{p}$), presents a fraction $(1-\corruptionprob)$ of questions with a relevant context. We call this training subset as the `retriever success' bucket.
For the remaining $\corruptionprob$ fraction of questions, only an irrelevant context (composed entirely of distractor documents) is provided. We call such a training subset as the `retriever failure' bucket.
However, this fine-tuning setup leads to conditional memorization bias, where the LLM either relies on relevant contexts or stores the necessary information in its parametric memory when presented with irrelevant contexts. As a result, the model may excel at handling specific retrieval scenarios but struggle to generalize effectively across varied contexts during inference. For some parts of the document, the LLM memorizes the content, while for others, it relies on the retrieved documents, resulting in inconsistent performance.

To address conditional memorization bias, we introduce \raftmix\ (Context-Augmented RAFT), which combines RAFT($\mathrm{0}$) and RAFT($\mathrm{1}$) with RAFT($\corruptionprob$) for more granular control over relevant and irrelevant contexts. In RAFT($\mathrm{0}$), all questions are paired with irrelevant contexts, teaching the model to rely solely on its internal memory. In RAFT($\mathrm{1}$), only relevant contexts are provided. By mixing RAFT($\mathrm{0}$), RAFT($\mathrm{1}$), and RAFT($\mathrm{p}$), \raftmix($\corruptionprob$) effectively mitigates Conditional Memorization Bias.

Our complete approach, \ourmethodshort\, significantly enhances \raftmix\ by incorporating paraphrased answer augmentation to address Canonical Answer Overfitting. 
While \raftmix\ ensures the model learns how to handle both relevant and irrelevant contexts without over-relying on one or the other, \ourmethodshort\ goes further by exploiting diverse paraphrased answers during training to effectively inject new knowledge. This augmentation ensures that the model does not overfit to a single canonical answer for each question, promoting deeper learning of underlying knowledge. 

Interestingly, answer augmentation implicitly takes care of conditional memorization bias as well.
When we randomly assign each training QA pair in $\mathcal{D}_{\text{aug.}}$ to `retriever success' or `retriever failure' bucket, it automatically distributes different QA pairs for the same question into different buckets.
Thus, when training with $\mathcal{D}_{\text{aug.}}$, we do not need context augmentation separately.

\subsection{Domain Identifiers}
\label{subsec:domain_indentifiers}
LLMs store vast amounts of information across different domains, which can lead to confusion when handling specialized questions. To ensure accurate and relevant responses, we introduce \emph{domain identifiers} that establish boundaries within the model’s parametric memory by specifying the context of each question. A domain identifier is a simple token or phrase prepended to each question during training and evaluation. For example:

\vspace{5pt}
\noindent\begin{minipage}{\textwidth}
	\noindent\texttt{"This question is from \{domain\_name\}.\\\{Question\}"\\}
\end{minipage} 
Here, the placeholder \texttt{\{domain\_name\}} is replaced with the specific domain (e.g., Healthcare, Legal), and \texttt{\{Question\}} is replaced with the actual question. This template ensures that each question is clearly tied to its domain, reducing ambiguity and helping the model differentiate between similar questions from different fields. Additionally, domain identifiers have minimal impact on general performance, as removing them may allow the model to return to its original state, preserving its generalization capabilities.

\subsection{Self-Selective Replay Buffer}
\label{subsec:self_selective_replay}
Typically, a replay buffer contains old samples from previous tasks on which the LLM has been trained. This helps the model retain general knowledge and mitigate catastrophic forgetting~\citep{zhang2024dissecting, ke2023continual, jang2021towards} while fine-tuning on new domain-specific data. However, since we do not have access to the original instruction-tuning data of the target LLM, 
we introduce a self-selective rehearsal replay buffer~\citep{gupta2024selective, huang2024mitigating}.

This approach differs from traditional replay buffers in that it generates new outputs for old inputs. 
Using the technique introduced in ~\citep{sudalairaj2024lab}, we generate diverse inputs that belong to different categories, such as code, math, reasoning, extraction, safety, writing \etc.  
We then pass each input (\textit{x}) through the LLM and use (\textit{x},LLM(\textit{x})) as an auxiliary dataset during the fine-tuning process.

By combining this self-selective replay buffer with the current synthetic QA data, we ensure that the model retains general knowledge while fine-tuning on domain-specific tasks, thereby mitigating catastrophic forgetting. 

\section{Experimental Setup}

\subsection{Objectives}
The main objectives of our experiments are to:
\begin{compactenum}[1)]
\item  demonstrate that \ourmethodshort\ successfully injects new knowledge in LLM while preserving its generic capabilities and show that it performs better than the other fine-tuning methods in both dimensions -- injecting new knowledge and preserving generic capabilities. 
\item demonstrate the importance of each of the components of \ourmethodshort, \viz, data identifier, replay-buffer, and answer multiplicity.
\item systematically analyze the issues with RAFT \citep{zhang2024raft}, \viz, contextual overfitting and canonical answer overfitting.
\end{compactenum}

All the code and test datasets used for our experiments are available on \href{https://github.com/kushagrabhushan/Systematic-Knowledge-Injection}{GitHub}. \footnote{\href{https://github.com/kushagrabhushan/Systematic-Knowledge-Injection}{https://github.com/kushagrabhushan/Systematic-Knowledge-Injection}} 

\subsection{Base model and Datasets}
\label{subsec:base_model_and_dataset}
We use Mistral-7B-Instruct-v0.2\footnote{\href{https://huggingface.co/mistralai/Mistral-7B-Instruct-v0.2}{mistralai/Mistral-7B-Instruct-v0.2}} as our base LLM and inject new knowledge in it using LoRA\cite{hu2022lora} tuning. 
To demonstrate the capability of \ourmethodshort\ to inject new knowledge, we would need information published after the cutoff date of the base model. Accordingly, we chose two Redbooks\footnote{Book 1: \href{https://www.redbooks.ibm.com/redpieces/pdfs/redp5736.pdf}{Do More with Less: Automating IBM Storage FlashSystem Tasks with REST APIs, Scripting, and Ansible}\\
Book 2: \href{https://www.redbooks.ibm.com/redpieces/pdfs/redp5711.pdf}{Red Hat OpenShift Container Platform on IBM Z and LinuxONE}
}
published on 23rd July and 16th May 2024 as our two different corpora. 
We use LlamaIndex\footnote{\href{https://www.llamaindex.ai/}{LlamaIndex}} to parse the PDFs into markdown and extract the text from the $5$ chapters of the first book and the $6$ chapters of the second book, respectively.

\noindent \textbf{Base data:}
We consider each chapter as a document and arbitrarily set $\maxtokens= 8000$ to split the chapters into chunks for creating synthetic QAs. 
Using base--data prompt $P_1$, we get a total of $5122$ and $33570$ QA pairs for book1 and book2, respectively, and we randomly split them into train/val/test splits. As a post-facto analysis, we compute the coverage of each chapter by the generated QA pairs and obtain $84.4\%$ and $82.5\%$ average coverage at the token level by the train data for book1 and book2, respectively. 

\noindent \textbf{Answer multiplicity in training data:}
We use multiplicity prompt $P_2$ to generate multiple answers for each question in the base dataset. We restrict the number of answers per question to $5$ and get $~4.6$ answers per question on average in the training data of both books. Generating multiple answers also results in an increase in the token-level coverage to 93.6\% and 92.4\%, respectively, for the two books. 

\noindent \textbf{Test Sets:} Along with the base data test set, we also created an additional factoid test set consisting of questions with up to eight words as answers. In the base data test set, answers often consist of multiple sentences and can be expressed in various paraphrased forms, making evaluation challenging. In contrast, factoid answers have a limited set of acceptable responses, making it easier to assess model performance. Evaluating models on this factoid test set provides a more direct measure of how many facts from the corpus have been injected into the model.
Additionally, we observed that several questions in the test sets contained phrases such as "in the above passage" and "in the given context." To address this, we removed such contextual questions by prompting a large language model (LLM) to filter them out. Table \ref{tab:filtering_prompt} provides the exact prompt for filtering the test set. All results in the paper are on the filtered base and factoid test set, with results on the factoid dataset presented in Appendix \ref{subsec:appendix_factoid}. Please see Table~\ref{tab:data_stats} in the Appendix~\ref{subsec:appendix_data_stats} for detailed statistics of all the training and test datasets.

\subsection{Training Details:}
Please see Appendix~\ref{subsec:appendix_training_details} for training details.

\subsection{Evaluation criteria}

We evaluate all the methods under the RAG setup, \ie, we fetch the top 5 passages that are similar to the question and provide them as context along with the question. 
To test under a realistic scenario, instead of setting up the index on passages from only book1 and book2, we  downloaded 34 red papers~\footnote{\url{https://www.redbooks.ibm.com/}} and indexed all of them. This results in the indexing of a total of 4,765 passages,
each with 512 tokens. We use LlamaIndex with BGE Embedding \cite{xiao2023c} to create the index. 
See Table~\ref{tab:ft_prompt} for the exact prompt used while fine-tuning. 

Following \cite{adlakha2023evaluating}, we use token level Recall \wrt\ the gold response as an automated metric to evaluate the generated response. 
In addition, we prompt Mixtral-8x22B-Instruct-v0.1\footnote{\href{https://huggingface.co/mistralai/Mixtral-8x22B-Instruct-v0.1}{mistralai/Mixtral-8x22B-Instruct-v0.1}} and LLaMA-3.3-70B-Instruct\footnote{\href{https://huggingface.co/meta-llama/Llama-3.3-70B-Instruct}{meta-llama/Llama-3.3-70B-Instruct}} to identify if the predicted answer conveys the same message as gold while answering the question or not. See Table~\ref{tab:llm_as_judge_prompt} for the prompt used. We report these metrics for the test split of base data.
To quantify catastrophic forgetting and model's generic reasoning capabilities, we use a diverse set of standard benchmarks, \viz, MMLU \cite{hendrycks2020measuring}, GSM8k \cite{cobbe2021gsm8k}, Hellaswag \cite{zellers2019hellaswag}, and, TruthfulQA \cite{lin2022truthfulqa}.

\subsection{Baselines}
\label{sec:baselines}
We compare \ourmethodshort\ against the following baselines. The first three are inspired by \citeauthor{zhang2024raft}, and the last one is an augmented version of RAFT.
\begin{compactenum}[1)]
\item Domain-specific fine-tuning (DSF): We train the base LLM using base data to generate a response to a question without accessing any retrieved passages. 
Since the QAs in the training data cover most of the corpus content, the LLM should be able to answer test queries, provided it has injected the knowledge shown during fine-tuning. 
See Table~\ref{tab:qaft_prompt} for the prompt used.
\item DSF+RAG: We prompt the DSF model with the top 5 retrieved passages along with the question. 
The prompt used here is the same as in the other baselines and \ourmethodshort\ (Table~\ref{tab:ft_prompt}).
\item RAFT($\corruptionprob*$): method proposed in \cite{zhang2024raft}, trained using base data. We treat corruption probability $\corruptionprob$ in RAFT as a hyper-parameter and select amongst $\{0.0, 0.2, 0.4, 0.6, 0.8, 1.0\}$ based on the validation loss. 
\item $\raftmixfn{\corruptionprob*}$: Here we augment the training data of RAFT($\corruptionprob$) with RAFT(0) and RAFT(1). As earlier, we treat corruption probability $\corruptionprob$ as a hyper-parameter and select the optimum $\corruptionprob*$ based on validation loss. 
\end{compactenum}

\section{Experimental Results}
\begin{table*}[]
\resizebox{\textwidth}{!}{
\begin{tabular}{@{}l|cccccccccc@{}}
\multicolumn{11}{l}{} \\ \midrule
& \multicolumn{3}{c|}{\textbf{Overall}} &
  \multicolumn{3}{c|}{\textbf{No Overlap}} &
  \multicolumn{3}{c|}{\textbf{Some Overlap}} &
  \multirow{2}{*}{\textbf{\begin{tabular}[c]{@{}c@{}}Reg. \\ Score\end{tabular}}} 
  \\ \cmidrule(r){1-10}
 &
  \textbf{Recall} &
  \textbf{Mixtral-J} &
  \multicolumn{1}{c|}{\textbf{LLaMA-J}} &
  \textbf{Recall} &
  \textbf{Mixtral-J} &
  \multicolumn{1}{c|}{\textbf{LLaMA-J}} &
  \textbf{Recall} &
  \textbf{Mixtral-J} &
  \multicolumn{1}{c|}{\textbf{LLaMA-J}} &
   \\ \midrule
\multicolumn{11}{c}{\textbf{Book 1}} \\ \midrule
\textbf{Base+RAG} &
  68.9 &
  80.7 &
  \multicolumn{1}{c|}{79.3} &
  57.7 &
  61.1 &
  \multicolumn{1}{c|}{56.4} &
  76.1 &
  93.4 &
  \multicolumn{1}{c|}{94.1} &
  56.4 \\
\textbf{DSF} &
  71.1 &
  84.0 &
  \multicolumn{1}{c|}{83.3} &
  72.3 &
  83.8 &
  \multicolumn{1}{c|}{81.9} &
  70.3 &
  84.1 &
  \multicolumn{1}{c|}{84.1} &
  -5.7 \\
\textbf{DSF+RAG} &
  68.1 &
  80.9 &
  \multicolumn{1}{c|}{82.7} &
  68.0 &
  71.9 &
  \multicolumn{1}{c|}{79.5} &
  68.2 &
  86.8 &
  \multicolumn{1}{c|}{84.8} &
  -5.7 \\
\textbf{RAFT} &
  70.6 &
  82.8 &
  \multicolumn{1}{c|}{84.0} &
  67.0 &
  76.0 &
  \multicolumn{1}{c|}{78.4} &
  73.0 &
  87.2 &
  \multicolumn{1}{c|}{87.6} &
  -5.4 \\
\textbf{CA-RAFT} &
  72.5 &
  85.2 &
  \multicolumn{1}{c|}{83.9} &
  69.2 &
  80.2 &
  \multicolumn{1}{c|}{78.3} &
  74.6 &
  88.4 &
  \multicolumn{1}{c|}{87.5} &
  -6.4 \\
\textbf{PA-RAG} &
  \textbf{77.6} &
  \textbf{95.3} &
  \multicolumn{1}{c|}{\textbf{93.6}} &
  \textbf{73.4} &
  \textbf{91.0} &
  \multicolumn{1}{c|}{\textbf{89.8}} &
  \textbf{80.3} &
  \textbf{98.1} &
  \multicolumn{1}{c|}{\textbf{96.1}} &
  \textbf{-1.2} \\ \midrule
\multicolumn{11}{c}{\textbf{Book 2}} \\ \midrule
\textbf{Base+RAG} &
  69.5 &
  67.8 &
  \multicolumn{1}{c|}{67.5} &
  67.1 &
  58.6 &
  \multicolumn{1}{c|}{58.9} &
  71.0 &
  73.2 &
  \multicolumn{1}{c|}{72.5} &
  56.4 \\
\textbf{DSF} &
  68.2 &
  70.3 &
  \multicolumn{1}{c|}{71.5} &
  68.1 &
  69.4 &
  \multicolumn{1}{c|}{70.7} &
  68.2 &
  70.7 &
  \multicolumn{1}{c|}{71.9} &
  -5.6 \\
\textbf{DSF+RAG} &
  66.0 &
  65.8 &
  \multicolumn{1}{c|}{66.3} &
  64.0 &
  61.7 &
  \multicolumn{1}{c|}{61.9} &
  67.2 &
  68.3 &
  \multicolumn{1}{c|}{68.9} &
  -5.6 \\
\textbf{RAFT} &
  70.9 &
  76.2 &
  \multicolumn{1}{c|}{77.2} &
  70.0 &
  73.1 &
  \multicolumn{1}{c|}{75.1} &
  71.4 &
  78.0 &
  \multicolumn{1}{c|}{78.5} &
  -4.8 \\
\textbf{CA-RAFT} &
  71.4 &
  78.2 &
  \multicolumn{1}{c|}{78.0} &
  70.2 &
  77.0 &
  \multicolumn{1}{c|}{74.8} &
  72.2 &
  78.9 &
  \multicolumn{1}{c|}{79.9} &
  -5.6 \\
\textbf{PA-RAG} &
  \textbf{76.4} &
  \textbf{86.2} &
  \multicolumn{1}{c|}{\textbf{85.7}} &
  \textbf{75.9} &
  \textbf{83.7} &
  \multicolumn{1}{c|}{\textbf{82.8}} &
  \textbf{76.6} &
  \textbf{87.7} &
  \multicolumn{1}{c|}{\textbf{87.4}} &
  \textbf{-1.8} \\ \bottomrule
\end{tabular}
}
\caption{Comparing \ourmethodshort\ with various baselines defined in Section~\ref{sec:baselines}. 
\textbf{Overall}: performance over the entire test set;
\textbf{No overlap}: the subset of test split where retriever fails;
\textbf{Some overlap}: subset where the retriever fetches at least one passage from the gold document;
\textbf{Reg. Score}: average performance on various benchmarks (GSM8k, Hellaswag, MMLU, and TruthfulQA). Mixtral-J and LLaMA-J stand for Mistral-8x22B-Instruct-v0.1 and Llama-3.3-70B-Instruct as judges, respectively. For the base model(Mistral-Instruct-v2), we show the average regression score, and for others, we show the drop from the base model. 
For RAFT, the optimal corruption probability, $\corruptionprob* = 0.6$ for both books;
For \raftmix, $\corruptionprob* = 0.2, 0.4$, and for \ourmethodshort\, $\corruptionprob* = 0.4, 0.6$, for book1 and book2, respectively.
}
\label{tab:main_table_breakup}
\end{table*}
\subsection{Comparison with baselines}
Table \ref{tab:main_table_breakup} compares the performance of our method with various baselines. 
\ourmethodshort\ performs the best in both dimensions: knowledge injection (measured by recall and LLM Judge), as well as maintaining the base model's general capabilities, as measured by the drop in average regression score. 
We observe that the base LLM performs better than most of the baselines when the retrieved passage overlaps with the gold document (we call it "some overlap" subset), demonstrating that the instruction-tuned Mistral has decent reading comprehension capabilities. 
However, when the retrieved passages do not overlap with the gold document ("No overlap" subset), the model has to recognize this and then answer from its memory.
We observe that all the baselines perform poorly in such a scenario.
On the other hand, \ourmethodshort\ beats all the baselines in both cases, demonstrating that:
1. When the retriever succeeds, \ourmethodshort\ can identify this and leverage the retrieved information to respond.
2. When the retriever fails, \ourmethodshort\ can ignore the retrieved information and answer correctly from its own parametric memory.

Next, we observe that \raftmix\ performs better than RAFT, confirming the presence of conditional memorization bias. Recall that such a bias occurs due to the static assignment of each question to retriever success or failure case during training. In \raftmix, when we augment RAFT($\corruptionprob$) with RAFT(0) and RAFT(1), we ensure that each question is seen with and without correct passages during training. This forces the model to learn that irrespective of the question, it has to leverage the information present in the context only when it is correct and rely on its own parametric knowledge when the retriever fails. 

Next, we observe that using a replay buffer, data identifier, and multiple answers for the same question significantly improves the model's performance without impacting its generic capabilities.
We note that \ourmethodshort\ implicitly takes care of conditional memorization bias. 
In Section~\ref{subsec:analysis_of_biases}, we empirically validate that this is indeed the case.

\begin{table*}[]
\begin{adjustbox}{max width=\textwidth}
\begin{tabular}{@{}lccccccccc|c@{}}
\toprule
 &
  \multicolumn{3}{c}{\textbf{Overall}} &
  \multicolumn{3}{c}{\textbf{No Overlap}} &
  \multicolumn{3}{c|}{\textbf{Some Overlap}} &
  \multicolumn{1}{l}{\multirow{2}{*}{\textbf{Reg. Scores}}} \\ \cmidrule(r){1-10}
 &
  \textbf{Recall} &
  \textbf{Mixtral-J} &
  \textbf{LLaMA-J} &
  \textbf{Recall} &
  \textbf{Mixtral-J} &
  \textbf{LLaMA-J} &
  \textbf{Recall} &
  \textbf{Mixtral-J} &
  \textbf{LLaMA-J} &
  \multicolumn{1}{l}{} \\ \midrule
\textbf{PA-RAG} & 77.6\% & 95.3\% & 93.6\% & 73.4\% & 91.0\% & 89.8\% & 80.3\% & 98.1\% & 96.1\% & 55.2 \\
- DI            & 76.5\% & 94.6\% & 94.8\% & 72.3\% & 92.2\% & 90.3\% & 79.1\% & 96.1\% & 97.6\% & -0.1 \\
- Replay Buffer & 77.0\% & 96.0\% & 94.8\% & 73.6\% & 92.8\% & 91.6\% & 79.2\% & 98.1\% & 96.9\% & -1.9 \\
- Multi Ans.    & 72.0\% & 84.9\% & 84.6\% & 65.8\% & 79.0\% & 78.9\% & 76.0\% & 88.8\% & 88.3\% & 0.6  \\ \bottomrule
\end{tabular}
\end{adjustbox}
\caption{Ablation results on book 1. The first row has absolute performance metrics of \ourmethodshort\ whereas the remaining rows show the change in the metric by removing the corresponding component from \ourmethodshort. DI: Data Identifier; Multi Ans: multiple answers. Mixtral-J and LLaMA-J stand for Mistral-8x22B-Instruct-v0.1 and Llama-3.3-70B-Instruct as judges, respectively.}
\label{tab:ablations}
\end{table*}

\subsection{Ablations} 
To understand the value addition by different components of \ourmethodshort, we run ablations by removing each of the components and retraining the model. 
Specifically, we ablate on replay buffer, Data Identifier (DI), and multiple answers (Multi Ans.).
Table \ref{tab:ablations} shows the results. For each ablation, we show the change in the metric \wrt\ \ourmethodshort.

We first observe that removing the replay buffer has a significant impact on the average regression score that measures the model's generic capabilities (a drop of 1.9 from \ourmethodshort\ and an overall drop of 3.1 from the base model).
Interestingly, removing the replay buffer negatively impacts the model's performance on the "some overlap" subset. 
This demonstrates that the replay buffer helps in bolstering the reading comprehension capabilities of the model as well.
On the other hand, performance on the "no overlap" subset improves slightly, demonstrating that the reduced burden of remembering existing skills helps in memorizing and recalling new information.

Next, we observe that removing the Data Identifier (DI) results in an overall drop in the model's performance.
The drop is more prominent over "no overlap" subset of the test data, demonstrating that the DI helps in injecting and recalling the domain knowledge when the model has to answer from its own parametric memory.

Finally, we observe that using multiple paraphrases of the same answer significantly impacts the performance -- recall drops by 5.65 pts from 77.0. As expected, dropping multiple answers has no negative impact on the model's generic capabilities.

\subsection{Analysis of conditional memorization bias}
\label{subsec:analysis_of_biases}

\begin{table}[]
\resizebox{\columnwidth}{!}{%
\begin{tabular}{@{}c|l|rrr@{}}
\toprule
\textbf{Ch.} &
  \textbf{Method} &
  \multicolumn{1}{c}{\textbf{Overall}} &
  \multicolumn{1}{c}{\textbf{\begin{tabular}[c]{@{}c@{}}No \\ overlap\end{tabular}}} &
  \multicolumn{1}{c}{\textbf{\begin{tabular}[c]{@{}c@{}}Some \\ overlap\end{tabular}}} \\ \midrule
\multirow{2}{*}{1-3} & \chaptercorruption & 74.1 & 71.3 & 76.0 \\
                     & \ourmethodshort    & 76.3 & 70.9 & 79.9 \\ \midrule
\multirow{2}{*}{4-5} & \chaptercorruption & 74.8 & 68.1 & 79.1 \\
                     & \ourmethodshort    & 78.4 & 75.0 & 80.5 \\ \bottomrule
\end{tabular}
}
\caption{Comparing chapter-wise assignment into `retriever success' and `retriever failure' buckets during training with \ourmethodshort\ that does random assignment.}
\label{tab:chapterwise_corruption}
\end{table}

In this section, we empirically demonstrate:
1). Conditional memorization bias exists, and it can significantly hamper the learning process, and
2). Answer multiplicity during training implicitly results in context augmentation when we randomly assign each augmented QA pair to either of the two buckets (`retriever success' or `retriever failure').

To do so, we systematically assign each QA pair to the `retriever success' and `retriever failure' buckets while training. 
In the first analysis, all QA pairs from chapters 1 to 3 from book 1 are assigned to the `retriever failure' bucket, and all QA pairs from chapters 4 and 5 are assigned to the `retriever success' bucket. 
Table \ref{tab:chapterwise_corruption} separately compares the token level recall for the test queries from chapters 1-3 and chapters 4-5. 
We observe that when there is no overlap between the retrieved passages and the gold documents, the performance over questions from chapters 4-5 is significantly poor for the chapter wise bucketing method.  
This happens because during training, all questions from chapters 4-5 had relevant context, and thus the model relied only on the given context to answer questions from chapters 4-5. Hence, when provided with irrelevant passages during testing, the model fails to recall that information and thus performs poorly.
We do not see any such pattern for questions from chapters 1-3.
Interestingly, when relevant information is provided during testing for questions from chapters 1-3, the model fails to leverage that, demonstrating that it learnt to always ignore the context when the question is from chapters 1-3.

Recall that our \ourmethodshort\ finetunes LLM using $\mathcal{D}_{aug.}$ that has multiple QA pairs for the same question. 
We show that in such a scenario, random assignment of train QA pairs to the `retriever success' or `retriever failure' bucket alleviates the need for explicit context augmentation in \ourmethodshort.   
To do so, we systematically assign all QA pairs with the same question to the same bucket. For a question, the `retriever failure' bucket is chosen with probability $\corruptionprob=0.4$.
In such a setup, we have systematically removed the context augmentation but have kept multiple answers.
Table \ref{tab:querwise_corruption} compares the token level recall of question-wise bucketing with \ourmethodshort\ that has QA-wise bucketing.
We observe that the overall token level recall worsens with question-wise bucketing. 
Notice that the gap is wider (2.5 pt) when there is no overlap between retrieved passages and the gold document for a test question. This demonstrates that the model struggles to either inject all the knowledge or recall it when required.
We attribute this to the lack of context augmentation in question-wise bucketing.

\begin{table}[]
\centering
\begin{tabular}{@{}l|ccc@{}}
\toprule
 &
  \textbf{Overall} &
  \textbf{\begin{tabular}[c]{@{}c@{}}No \\ overlap\end{tabular}} &
  \textbf{\begin{tabular}[c]{@{}c@{}}Some\\ overlap\end{tabular}} \\ \midrule
\querywisecorruption &
  76.2 &
  71.1 &
  79.5 \\
\ourmethodshort &
  77.6 &
  73.4 &
  80.3 \\ \bottomrule
\end{tabular}
\caption{Comparing question-wise assignment into `retriever success' and `retriever failure' buckets during training with \ourmethodshort\ that randomly assigns QA pairs.}
\label{tab:querwise_corruption}
\end{table}

\subsection{Effect of model size and model family}
All the results presented above are obtained by fine-tuning a Mistral-7B model.
To evaluate the robustness of our method to variations in model size and architecture, we train two additional models from the LLaMA family, LLaMA-2-7B-Chat\footnote{\href{https://huggingface.co/meta-llama/Llama-2-7b-chat-hf}{meta-llama/Llama-2-7b-chat-hf}} and LLaMA-2-13B-Chat\footnote{\href{https://huggingface.co/meta-llama/Llama-2-13b-chat-hf}{meta-llama/Llama-2-13b-chat-hf}}.
Table~\ref{tab:main_table_llama7b} reports our findings.
In both cases, we observe trends similar to Mistral-7B, demonstrating that \ourmethodshort\ is robust to varying architectures and model sizes.

\subsection{Effect of ingesting already seen knowledge}

\label{subsec:clapnq}
We conduct additional experiments using a small subset (50 documents and 50 queries) of the recently proposed CLAPNQ dataset \citep{rosenthal2025clapnq}.
It is derived from NQ dataset \citep{kwiatkowski2019natural} and consists of a human-annotated corpus of actual user queries and corresponding answers from Wikipedia articles. 
Mistral may not have seen the specific QA pairs during training, but it likely encountered the underlying information from Wikipedia pages in various paraphrased forms.
As a result, there is no new knowledge to be ingested, and we do not expect our method to yield significant performance gains over the base model.
See table~\ref{tab:clapnq} for the results. 

Surprisingly, RAFT performs worse than the base model, while \ourmethodshort's performance remains at par with the base model. 
We hypothesize two causes for this: 
First, RAFT is trained using only a single answer per question, which may result in overfitting to canonical answers and losing its ability to recall knowledge already seen during pretraining. 
In contrast, \ourmethodshort\ trains on multiple answers and varied contexts, preserving and leveraging this ability.
Second, our synthetically generated QAs cover only 49\% of a document on average. However, it is plausible that the base model would have seen the remaining 51\% during pretraining, giving it an edge. RAFT results in overfitting to 49\% of the information, whereas \ourmethodshort\ generalises better by retaining the pre-existing knowledge and skills due to the replay buffer.

\section{Conclusion}\label{sec:conclusion}
We introduced \ourmethodshort, a novel framework designed to enhance the fine-tuning process for domain-specific RAG tasks. By incorporating context augmentation and answer multiplicity through paraphrasing,
our approach effectively mitigates both conditional memorization bias and canonical answer overfitting. 
It results in a more adaptable LLM that is robust to retriever errors on domain-specific questions.
We use a novel replay buffer technique along with a data identifier that mitigates catastrophic forgetting.
Our experimental results demonstrate that \ourmethodshort\ outperforms baseline methods, offering a promising direction for improving domain-specific knowledge injection in LLMs.
\section{Limitations} While \ourmethodshort\ demonstrates promising results in reducing catastrophic forgetting, it does not completely eliminate the issue. The retained parametric knowledge may still degrade when new domain-specific information is introduced due to domain-specific fine-tuning. Additionally, the effectiveness of \ourmethodshort\ depends on a strong LLM to generate high-quality QA pairs from domain documents. If the LLM struggles to generate accurate or contextually rich pairs, the overall performance may be affected. Moreover, the enhancements introduced by \ourmethodshort\, such as paraphrased answer augmentation and dynamic retrieval simulation, require additional computational resources, which can be a concern when injecting large-scale domain data. Despite these limitations, \ourmethodshort\ shows considerable promise, though further work is needed to address these challenges and make it more robust and widely applicable.
\section*{Acknowledgements}
We thank anonymous reviewers for
their insightful comments and suggestions that helped in further improving our paper.
We also thank Meghanadh Pulivarthi for reviewing our code independently and Ojas Gramopadhye for helping out with many teething issues during the experimental setup.

\bibliography{custom}

\newpage
\clearpage 
\appendix
\onecolumn 
\section{Experimental Details}
\label{sec_appendix_expt}

\subsection{Data statistics}
\label{subsec:appendix_data_stats}

See Table \ref{tab:data_stats} for detailed statistics of all the training and test datasets used in our experiments.

\begin{table*}[h]
\centering
\begin{tabular}{@{}lcccccccc@{}}
\toprule
 &
  \textbf{\begin{tabular}[c]{@{}c@{}}\#Chapters/\\ Pages\end{tabular}} &
  \textbf{\#Trn} &
  \textbf{\#Val} &
  \textbf{\begin{tabular}[c]{@{}c@{}}Test \\ (Raw)\end{tabular}} &
  \textbf{\begin{tabular}[c]{@{}c@{}}Test \\ (Filtered)\end{tabular}} &
  \textbf{\begin{tabular}[c]{@{}c@{}}\%cover. \\ trn\end{tabular}} &
  \textbf{\begin{tabular}[c]{@{}c@{}}\# Aug. \\ trn.\end{tabular}} &
  \textbf{\begin{tabular}[c]{@{}c@{}}\%cover.\\ aug.\end{tabular}} \\ \midrule
\textbf{Book1} &
  5 / 76 &
  4,122 &
  515 &
  515 &
  425 &
  84.4 &
  18,986 &
  93.6 \\
\textbf{Book2} &
  6 / 158 &
  27,056 &
  3,382 &
  3,132 &
  2,269 &
  82.5 &
  126,213 &
  92.4 \\ \bottomrule
\end{tabular}
\caption{Data statistics. 
\#Trn: number of training QA pairs in base data;
\#Val: number of validation QA pairs.
\#Test (Raw): number of test QA pairs before filtering.
\#Test (Filtered): number of test QA pairs after filtering.
\%cover. trn: average coverage of the chapters by the train QA pairs in the base data. It is computed as token level overlap between the chapter and all QA pairs from that chapter.
\#Aug. trn.: number of training QA pairs in the augmented train data.
\%cover. aug.: average coverage of the chapters by the train QA pairs in the augmented base data.
}
\label{tab:data_stats}
\end{table*}

\subsection{\ourmethodshort\ training details}
\label{subsec:appendix_training_details}
We use Huggingface's SFTTrainer\footnote{\href{https://huggingface.co/docs/trl/sft_trainer}{docs/trl/sft\_trainer}}
to finetune the base model separately for book1 and book2 using LoRA adapters with rank 16 and 32, respectively.
We experimented with the learning rates of $1e-5$ and $1e-4$ and selected $1e-4$ based on validation set performance. 
We train for 400 and 1200 steps for book1 and book2, respectively, with an effective batch size of 256 (16 gradient accumulation steps on 4 A100 80GB GPUs with 4 batch-size on each GPU) and select the best checkpoint based on validation loss. Training time for the two models is less than 5 and 15 hours, respectively.

\subsection{Effect of domain identifiers and replay buffer on catastrophic forgetting}
\label{subsec:appendix_regression_scores}
\begin{table}[h]
\centering
\resizebox{\columnwidth}{!}{%
\begin{tabular}{@{}c|cc|c|c|cc|c@{}}
\toprule
\multicolumn{1}{r|}{\textbf{}} &
  \multicolumn{2}{c|}{\textbf{GSM8k}} &
  \textbf{HellaSwag} &
  \textbf{MMLU} &
  \multicolumn{2}{c|}{\textbf{TQA}} &
  \textbf{TQA\_Gen} \\ \midrule
\multicolumn{1}{r|}{\textbf{}} &
  \textbf{Flexible} &
  \textbf{Strict} &
  \textbf{Acc} &
  \textbf{Acc} &
  \textbf{\begin{tabular}[c]{@{}c@{}}TQA\_MC1 \\ (acc)\end{tabular}} & 
  \textbf{\begin{tabular}[c]{@{}c@{}}TQA\_MC2 \\ (acc)\end{tabular}} &   
  \textbf{\begin{tabular}[c]{@{}c@{}}RougeL \\ (acc)\end{tabular}} \\
\multicolumn{1}{l|}{\textbf{Base Model}}                           & 41.9 & 41.6    & 66.0    & 59.0    & 52.4    & 66.8    & 54.1 \\
\multicolumn{1}{l|}{\textbf{DSF}}                                  & 35.7                      & 35.4    & 63.9    & 57.6    & 39.0    & 56.1    & 48.7                      \\
\multicolumn{1}{l|}{\textbf{DSF + DI}}                             & 37.1                      & 36.8    & 64.2    & 57.7    & 39.8    & 57.0    & 49.3                      \\
\multicolumn{1}{l|}{\textbf{DSF + RB}}                           & 42.8                      & 42.3    & 64.8    & 59.2    & 47.7    & 65.2    & 55.2                      \\
\multicolumn{1}{l|}{\textbf{DSF + DI + RB}}                        & 41.7                      & 41.2    & 64.8    & 58.7    & 48.1    & 65.7    & 54.5         \\ \bottomrule
\end{tabular}%
}
\caption{Regression Scores For Book 1: Performance of the training methods discussed in paper on various benchmark tasks using Mistral-7B-v0.1. \textbf{Flexible, Strict} are the different criteria for assessing the exact match metric. \textbf{TQA} refers to the TruthfulQA benchmark, where \textbf{MC1} is the subset of multiple choice question with one answer and MC2 is the subset of questions with more than one answer}
\label{tab:reg-book-1}
\end{table}

\begin{table*}[]
\centering
\resizebox{\textwidth}{!}{%
\begin{tabular}{@{}c|cc|c|c|cc|c@{}}
\toprule
\multicolumn{1}{r|}{\textbf{}} &
  \multicolumn{2}{c|}{\textbf{GSM8k}} &
  \textbf{HellaSwag} &
  \textbf{MMLU} &
  \multicolumn{2}{c|}{\textbf{TQA}} &
  \textbf{TQA\_Gen} \\ \midrule
\multicolumn{1}{r|}{\textbf{}} &
  \textbf{Flexible} &
  \textbf{Strict} &
  \textbf{Acc} &
  \textbf{Acc} &
  \textbf{\begin{tabular}[c]{@{}c@{}}TQA\_MC1 \\ (acc)\end{tabular}} & 
  \textbf{\begin{tabular}[c]{@{}c@{}}TQA\_MC2 \\ (acc)\end{tabular}} &   
  \textbf{\begin{tabular}[c]{@{}c@{}}RougeL \\ (acc)\end{tabular}} \\
\multicolumn{1}{l|}{\textbf{Base Model}} &
  \multicolumn{1}{c}{41.9} &
  \multicolumn{1}{c|}{41.6} &
  \multicolumn{1}{c|}{66.0} &
  \multicolumn{1}{c|}{59.0} &
  \multicolumn{1}{c}{52.4} &
  \multicolumn{1}{c|}{66.8} &
  54.1 \\
\multicolumn{1}{l|}{\textbf{DSF}} &
  37.6 &
  \multicolumn{1}{c|}{37.3} &
  \multicolumn{1}{c|}{62.8} &
  \multicolumn{1}{c|}{55.6} &
  37.6 &
  \multicolumn{1}{c|}{55.1} &
  51.7 \\
\multicolumn{1}{l|}{\textbf{DSF + DI}} &
  36.5 &
  \multicolumn{1}{c|}{36.3} &
  \multicolumn{1}{c|}{62.8} &
  \multicolumn{1}{c|}{56.5} &
  40.4 &
  \multicolumn{1}{c|}{57.8} &
  53.0 \\
\multicolumn{1}{l|}{\textbf{DSF + RB}} &
  41.0 &
  \multicolumn{1}{c|}{40.7} &
  \multicolumn{1}{c|}{64.1} &
  \multicolumn{1}{c|}{58.9} &
  48.3 &
  \multicolumn{1}{c|}{64.5} &
  54.0 \\
\multicolumn{1}{l|}{\textbf{DSF + DI + RB}} &
  40.7 &
  \multicolumn{1}{c|}{40.4} &
  \multicolumn{1}{c|}{64.2} &
  \multicolumn{1}{c|}{58.4} &
  48.7 &
  \multicolumn{1}{c|}{65.0} &
  53.2 \\
 \bottomrule
\end{tabular}%
}
\caption{Regression Scores for Book 2: Performance of the training methods discussed in paper on various benchmark tasks using Mistral-7B-v0.1. \textbf{Flexible, Strict} are the different criteria for assessing the exact match metric. \textbf{TQA} refers to the TruthfulQA benchmark, where \textbf{MC1} is the subset of multiple choice question with one answer and MC2 is the subset of questions with more than one answer}
\label{tab:reg-book-2}
\end{table*}

Here, we demonstrate the effectiveness of both Domain Identifiers and Replay Buffer to mitigate catastrophic forgetting. 
Tables~\ref{tab:reg-book-1} and \ref{tab:reg-book-2} report the individual regression scores for each task. 
We use github repo\cite{eval-harness} to compute all the scores.
For GSM8k, we take the average of Flexible and Strict Match.
For TruthfulQA (TQA), we take the average accuracy over MC1 and MC2.
We consider TruthfulQA\_Gen separately from MC1 and MC2 and use RougeL to quantify the performance.
Reported average score is the average of these five scores.

\section{Experimental Analysis of LLaMA Models}

Tables \ref{tab:main_table_llama7b} and \ref{tab:main_table_llama_13b} show the results of , LLaMA-2-7B-Chat and , LLaMA-2-13B-Chat respectively on both book 1 and book 2. We see in both the cases that \ourmethodshort  \hspace{0.1cm} outperforms all baselines, proving its robustness to model architecture and size. 

\begin{table*}[]
\resizebox{\textwidth}{!}{
\begin{tabular}{@{}lcccccccccc@{}}
\toprule
\multicolumn{4}{c|}{\textbf{Overall}} &
  \multicolumn{3}{c|}{\textbf{No Overlap}} &
  \multicolumn{3}{c|}{\textbf{Some Overlap}} &
  \multicolumn{1}{l}{\multirow{2}{*}{\textbf{Reg. Scores}}} \\ \cmidrule(r){1-10}
 &
  \textbf{Recall} &
  \textbf{Mixtral-J} &
  \multicolumn{1}{c|}{\textbf{LLaMA-J}} &
  \textbf{Recall} &
  \textbf{Mixtral-J} &
  \multicolumn{1}{c|}{\textbf{LLaMA-J}} &
  \textbf{Recall} &
  \textbf{Mixtral-J} &
  \multicolumn{1}{c|}{\textbf{LLaMA-J}} &
  \multicolumn{1}{l}{} \\ \midrule
\multicolumn{11}{c}{\textbf{Book 1}} \\ \midrule
\textbf{Base+RAG} &
  69.7 &
  73.2 &
  70.9 &
  60.7 &
  53.3 &
  48.8 &
  75.5 &
  86.0 &
  \multicolumn{1}{c|}{84.9} &
  41.9 \\
\textbf{DSF} &
  66.0 &
  72.9 &
  74.2 &
  67.0 &
  70.7 &
  73.5 &
  65.3 &
  74.4 &
  \multicolumn{1}{c|}{74.7} &
  -3.1 \\
\textbf{DSF+RAG} &
  57.6 &
  64.9 &
  68.8 &
  56.6 &
  59.3 &
  64.8 &
  58.3 &
  68.6 &
  \multicolumn{1}{c|}{71.4} &
  -3.1 \\
\textbf{RAFT} &
  68.9 &
  77.2 &
  78.5 &
  65.4 &
  68.3 &
  69.3 &
  71.2 &
  82.9 &
  \multicolumn{1}{c|}{84.4} &
  -1.9 \\
\textbf{CA-RAFT} &
  71.8 &
  79.5 &
  78.8 &
  67.1 &
  70.7 &
  70.7 &
  74.8 &
  85.3 &
  \multicolumn{1}{c|}{84.0} &
  -2.9 \\
\textbf{PA-RAG} &
  \textbf{73.7} &
  \textbf{89.6} &
  \textbf{88.7} &
  \textbf{69.7} &
  \textbf{83.8} &
  \textbf{83.8} &
  \textbf{76.3} &
  \textbf{93.4} &
  \multicolumn{1}{c|}{\textbf{91.8}} &
  -0.2 \\ \midrule
\multicolumn{11}{c}{\textbf{Book 2}} \\ \midrule
\textbf{Base+RAG} &
  70.9 &
  60.2 &
  57.9 &
  68.0 &
  50.7 &
  49.6 &
  72.7 &
  65.7 &
  \multicolumn{1}{c|}{62.8} &
  41.9 \\
\textbf{DSF} &
  67.7 &
  68.0 &
  69.0 &
  67.4 &
  66.1 &
  66.2 &
  67.9 &
  69.1 &
  \multicolumn{1}{c|}{70.7} &
  -2.9 \\
\textbf{DSF+RAG} &
  55.3 &
  54.3 &
  55.9 &
  55.1 &
  49.6 &
  52.8 &
  55.4 &
  57.0 &
  \multicolumn{1}{c|}{57.7} &
  -2.9 \\
\textbf{RAFT} &
  69.0 &
  70.5 &
  71.1 &
  68.0 &
  66.5 &
  68.2 &
  69.6 &
  72.8 &
  \multicolumn{1}{c|}{72.8} &
  -2.9 \\
\textbf{CA-RAFT} &
  70.5 &
  72.7 &
  73.1 &
  69.9 &
  70.8 &
  70.3 &
  70.9 &
  73.8 &
  \multicolumn{1}{c|}{74.7} &
  -4.2 \\
\textbf{PA-RAG} &
  \textbf{73.7} &
  \textbf{82.4} &
  \textbf{82.0} &
  \textbf{73.2} &
  \textbf{78.6} &
  \textbf{78.1} &
  \textbf{74.1} &
  \textbf{84.7} &
  \multicolumn{1}{c|}{\textbf{84.3}} &
  -0.7 \\ \bottomrule
\end{tabular}
}
\caption{Comparison of , LLaMA-2-7B-Chat trained using \ourmethodshort\ against baselines mentioned in \ref{sec:baselines}.}
\label{tab:main_table_llama7b}
\end{table*}

\begin{table*}[]
\resizebox{\textwidth}{!}{
\begin{tabular}{@{}lcccccccccc@{}}
\toprule
\multicolumn{4}{c}{\textbf{Overall}} &
  \multicolumn{3}{c}{\textbf{No Overlap}} &
  \multicolumn{3}{c|}{\textbf{Some Overlap}} &
  \multirow{2}{*}{\textbf{Reg. Scores}} \\ \cmidrule(r){1-10}
 &
  \textbf{Recall} &
  \textbf{Mixtral-J} &
  \textbf{LLaMA-J} &
  \textbf{Recall} &
  \textbf{Mixtral-J} &
  \textbf{LLaMA-J} &
  \textbf{Recall} &
  \textbf{Mixtral-J} &
  \multicolumn{1}{c|}{\textbf{LLaMA-J}} &
   \\ \midrule
\multicolumn{11}{c}{\textbf{Book 1}}                                                                         \\ \midrule
\textbf{Base+RAG} & 70.8   & 74.4   & 73.5 & 61.6 & 54.5 & 54.2 & 76.7 & 87.2 & \multicolumn{1}{c|}{86.0} & 45.4 \\
\textbf{DSF}      & 70.2   & 78.6   & 78.9 & 70.2 & 74.3 & 73.7 & 70.2 & 81.4 & \multicolumn{1}{c|}{82.4} & -1.9 \\
\textbf{DSF+RAG}  & 58.3   & 65.2   & 71.4 & 56.6 & 58.1 & 66.5 & 59.3 & 69.8 & \multicolumn{1}{c|}{74.6} & -1.9 \\
\textbf{RAFT}     & 70.3   & 79.3   & 79.4 & 66.0 & 69.5 & 71.7 & 73.1 & 85.7 & \multicolumn{1}{c|}{84.4} & -1.1 \\
\textbf{CA-RAFT}  & 72.7   & 83.5   & 83.6 & 67.5 & 74.3 & 77.0 & 76.1 & 89.5 & \multicolumn{1}{c|}{87.8} & -1.9 \\
\textbf{PA-RAG} &
  \textbf{77.1} &
  \textbf{92.0} &
  \textbf{92.0} &
  \textbf{73.6} &
  \textbf{89.2} &
  \textbf{89.8} &
  \textbf{79.4} &
  \textbf{93.8} &
  \multicolumn{1}{c|}{\textbf{93.4}} &
  -1.8 \\ \midrule
\multicolumn{11}{c}{\textbf{Book 2}}                                                                         \\ \midrule
\textbf{Base+RAG} & 72.9 & 63.1 & 62.4 & 70.2 & 53.9 & 53.5 & 74.4 & 68.4 & \multicolumn{1}{c|}{67.6} & 45.4 \\
\textbf{DSF}      & 67.5 & 68.8 & 69.9 & 67.7 & 68.7 & 69.7 & 67.4 & 68.9 & \multicolumn{1}{c|}{70.1} & -1.9 \\
\textbf{DSF+RAG}  & 59.3 & 55.7 & 59.8 & 58.6 & 52.2 & 56.6 & 59.7 & 57.8 & \multicolumn{1}{c|}{61.6} & -1.9 \\
\textbf{RAFT}     & 68.7 & 71.4 & 72.7 & 67.7 & 67.9 & 69.7 & 69.3 & 73.5 & \multicolumn{1}{c|}{74.4} & -1.5 \\
\textbf{CA-RAFT}  & 68.8 & 72.6 & 73.3 & 67.4 & 68.9 & 68.9 & 69.6 & 74.8 & \multicolumn{1}{c|}{75.9} & -2.3 \\
\textbf{PA-RAG} &
  \textbf{74.7} &
  \textbf{83.6} &
  \textbf{83.3} &
  \textbf{74.0} &
  \textbf{81.0} &
  \textbf{80.4} &
  \textbf{75.1} &
  \textbf{85.2} &
  \multicolumn{1}{c|}{\textbf{85.0}} &
  -0.3 \\ \bottomrule
\end{tabular}
}
\caption{Comparison of , LLaMA-2-13B-Chat trained using \ourmethodshort \hspace{0.1cm}against baselines mentioned in \ref{sec:baselines}.}
\label{tab:main_table_llama_13b}
\end{table*}
\newpage
\section{Results on Factoid Dataset}
\label{subsec:appendix_factoid}

As mentioned in \ref{subsec:base_model_and_dataset}, we curated a factoid dataset consisting of QA pairs containing only factoid answers, i.e., short, factual questions. Tables \ref{tab:factoid_mistral}, \ref{tab:factoid_llama7b} \& \ref{tab:factoid_llama13b} show the comparisons between \ourmethodshort  \hspace{0.1cm} and various other baselines, on all three models, Mistral-7B-v0.1 , LLaMA-2-7b-chat-hf  and LLaMA-2-13b-chat-hf. 

\begin{table*}[]
\resizebox{\textwidth}{!}{
\begin{tabular}{@{}lccccccccc@{}}
\toprule
\multicolumn{4}{c}{\textbf{Overall}} &
  \multicolumn{3}{c}{\textbf{No Overlap}} &
  \multicolumn{3}{c}{\textbf{Some Overlap}} \\ \midrule
\multicolumn{10}{c}{\textbf{Book 1}}                                                                        \\ \midrule
 &
  \textbf{Recall} &
  \textbf{Mixtral-J} &
  \textbf{LLaMA-J} &
  \textbf{Recall} &
  \textbf{Mixtral-J} &
  \textbf{LLaMA-J} &
  \textbf{Recall} &
  \textbf{Mixtral-J} &
  \textbf{LLaMA-J} \\
\textbf{Base+RAG} & 56.1 & 73.0 & 69.7 & 30.1 & 50.0 & 44.0 & 76.1 & 93.4 & 87.6          \\
\textbf{DSF}      & 69.2 & 83.4 & 81.0 & 66.4 & 79.9 & 77.5 & 70.3 & 84.1 & 83.4          \\
\textbf{DSF+RAG}  & 66.1 & 80.7 & 79.5 & 56.0 & 71.6 & 67.9 & 68.2 & 86.8 & 87.6          \\
\textbf{RAFT}     & 70.3 & 84.1 & 84.1 & 62.1 & 75.7 & 75.0 & 80.3 & 98.1 & 90.5          \\
\textbf{CA-RAFT}  & 72.5 & 85.6 & 85.4 & 64.4 & 74.6 & 76.3 & 73.0 & 87.2 & \textbf{91.7} \\
\textbf{PA-RAG} &
  \textbf{76.7} &
  \textbf{88.0} &
  \textbf{87.3} &
  \textbf{71.3} &
  \textbf{81.7} &
  \textbf{81.7} &
  \textbf{74.6} &
  \textbf{88.4} &
  91.3 \\ \midrule
\multicolumn{10}{c}{\textbf{Book 2}}                                                                        \\ \midrule
\textbf{Base+RAG} & 56.3 & 61.8 & 60.5 & 46.3 & 47.2 & 46.4 & 62.6 & 70.9 & 69.4          \\
\textbf{DSF}      & 53.9 & 62.5 & 60.1 & 51.4 & 58.6 & 56.4 & 55.5 & 64.9 & 62.5          \\
\textbf{DSF+RAG}  & 57.5 & 66.1 & 65.7 & 51.2 & 57.9 & 57.3 & 61.3 & 71.2 & 70.9          \\
\textbf{RAFT}     & 61.5 & 71.6 & 70.7 & 55.4 & 64.6 & 63.9 & 65.3 & 76.0 & 75.0          \\
\textbf{CA-RAFT}  & 62.2 & 72.1 & 71.1 & 56.9 & 65.4 & 64.1 & 65.6 & 76.3 & 75.5          \\
\textbf{PA-RAG} &
  \textbf{68.5} &
  \textbf{80.3} &
  \textbf{79.2} &
  \textbf{64.8} &
  \textbf{75.2} &
  \textbf{73.9} &
  \textbf{70.8} &
  \textbf{83.4} &
  \textbf{82.6} \\ \bottomrule
\end{tabular}
}
\caption{Results of \ourmethodshort \hspace{0.1 pt} and other baselines using  Mistral-7B-v0.1 on the factoid dataset}
\label{tab:factoid_mistral}
\end{table*}

\begin{table*}[]
\resizebox{\textwidth}{!}{
\begin{tabular}{@{}lccccccccc@{}}
\toprule
\multicolumn{4}{c|}{\textbf{Overall}} &
  \multicolumn{3}{c|}{\textbf{No Overlap}} &
  \multicolumn{3}{c}{\textbf{Some Overlap}} \\ \midrule
\multicolumn{10}{c}{\textbf{Book 1}} \\ \midrule
 &
  \textbf{Recall} &
  \textbf{Mixtral-J} &
  \textbf{LLaMA-J} &
  \textbf{Recall} &
  \textbf{Mixtral-J} &
  \textbf{LLaMA-J} &
  \textbf{Recall} &
  \textbf{Mixtral-J} &
  \textbf{LLaMA-J} \\
\textbf{Base+RAG} &
  52.5 &
  65.6 &
  60.7 &
  26.2 &
  37.3 &
  30.8 &
  71.0 &
  85.5 &
  81.7 \\
\textbf{DSF} &
  61.5 &
  73.2 &
  73.2 &
  61.0 &
  73.4 &
  71.6 &
  61.9 &
  73.0 &
  74.3 \\
\textbf{DSF+RAG} &
  53.0 &
  68.5 &
  68.5 &
  46.0 &
  62.7 &
  60.7 &
  57.9 &
  72.6 &
  73.9 \\
\textbf{RAFT} &
  65.0 &
  79.8 &
  78.8 &
  55.4 &
  70.4 &
  70.4 &
  71.8 &
  86.3 &
  84.6 \\
\textbf{CA-RAFT} &
  69.6 &
  82.7 &
  81.0 &
  \textbf{65.0} &
  76.9 &
  75.1 &
  72.8 &
  86.7 &
  85.1 \\
\textbf{PA-RAG} &
  \textbf{71.4} &
  \textbf{86.8} &
  \textbf{86.3} &
  62.3 &
  \textbf{79.3} &
  \textbf{79.8} &
  \textbf{77.8} &
  \textbf{92.1} &
  \textbf{90.9} \\ \midrule
\multicolumn{10}{c}{\textbf{Book 2}} \\ \midrule
\textbf{Base+RAG} &
  53.6 &
  58.5 &
  56.3 &
  43.1 &
  42.9 &
  41.2 &
  60.1 &
  68.3 &
  65.8 \\
\textbf{DSF} &
  51.5 &
  58.5 &
  57.1 &
  46.8 &
  52.7 &
  51.2 &
  54.4 &
  62.2 &
  60.7 \\
\textbf{DSF+RAG} &
  50.0 &
  57.1 &
  57.0 &
  42.9 &
  46.6 &
  46.5 &
  54.4 &
  63.7 &
  63.5 \\
\textbf{RAFT} &
  58.3 &
  67.1 &
  65.4 &
  51.0 &
  57.4 &
  55.2 &
  62.9 &
  73.3 &
  71.8 \\
\textbf{CA-RAFT} &
  59.2 &
  68.2 &
  66.6 &
  54.0 &
  62.1 &
  59.7 &
  62.5 &
  72.0 &
  70.9 \\
\textbf{PA-RAG} &
  \textbf{62.8} &
  \textbf{73.8} &
  \textbf{73.0} &
  \textbf{57.2} &
  \textbf{65.0} &
  \textbf{65.2} &
  \textbf{66.2} &
  \textbf{79.3} &
  \textbf{77.9} \\ \bottomrule
\end{tabular}
}
\caption{Results of \ourmethodshort \hspace{0.1cm} and other baselines using  , LLaMA-2-7B-Chat on the factoid dataset}
\label{tab:factoid_llama7b}
\end{table*}

\begin{table*}[]
\center
\resizebox{\textwidth}{!}{
\begin{tabular}{@{}lccccccccc@{}}
\toprule
\multicolumn{4}{c|}{\textbf{Overall}} &
  \multicolumn{3}{c|}{\textbf{No Overlap}} &
  \multicolumn{3}{c}{\textbf{Some Overlap}} \\ \midrule
\multicolumn{10}{c}{\textbf{Book 1}}                                                               \\ \midrule
 &
  \textbf{Recall} &
  \textbf{Mixtral-J} &
  \textbf{LLaMA-J} &
  \textbf{Recall} &
  \textbf{Mixtral-J} &
  \textbf{LLaMA-J} &
  \textbf{Recall} &
  \textbf{Mixtral-J} &
  \textbf{LLaMA-J} \\
\textbf{Base+RAG} & 54.1   & 65.8   & 62.7 & 29.1 & 41.6 & 36.1 & 71.7 & 82.5 & 81.3 \\
\textbf{DSF}      & 67.9   & 80.7   & 79.2 & 67.3 & 77.5 & 75.1 & 68.3 & 83.0 & 82.1 \\
\textbf{DSF+RAG}  & 55.5   & 69.0   & 69.3 & 43.6 & 58.6 & 60.4 & 63.8 & 76.3 & 75.5 \\
\textbf{RAFT}     & 66.9   & 81.0   & 81.7 & 63.7 & 73.4 & 75.1 & 69.1 & 86.3 & 86.3 \\
\textbf{CA-RAFT}  & 70.3   & 82.7   & 82.2 & 67.8 & 78.7 & 78.1 & 72.0 & 85.5 & 85.1 \\
\textbf{PA-RAG} &
  \textbf{74.9} &
  \textbf{86.1} &
  \textbf{86.8} &
  \textbf{68.1} &
  \textbf{78.7} &
  \textbf{79.8} &
  \textbf{79.7} &
  \textbf{91.3} &
  \textbf{91.7} \\ \midrule
\multicolumn{10}{c}{\textbf{Book 2}}                                                               \\ \midrule
\textbf{Base+RAG} & 56.3 & 59.8 & 57.3 & 45.5 & 44.3 & 42.4 & 63.0 & 69.5 & 66.7 \\
\textbf{DSF}      & 52.9 & 61.0 & 59.4 & 49.5 & 54.9 & 53.5 & 55.0 & 64.9 & 63.2 \\
\textbf{DSF+RAG}  & 47.0 & 51.7 & 54.5 & 41.0 & 43.6 & 46.9 & 50.8 & 56.8 & 59.3 \\
\textbf{RAFT}     & 58.7 & 67.9 & 66.7 & 52.0 & 58.2 & 57.1 & 63.0 & 74.0 & 72.8 \\
\textbf{CA-RAFT}  & 59.1 & 68.2 & 66.8 & 52.7 & 59.2 & 57.8 & 63.1 & 73.9 & 72.5 \\
\textbf{PA-RAG} &
  \textbf{66.5} &
  \textbf{77.7} &
  \textbf{77.0} &
  \textbf{61.5} &
  \textbf{69.9} &
  \textbf{69.7} &
  \textbf{69.7} &
  \textbf{82.6} &
  \textbf{81.6} \\ \bottomrule
\end{tabular}
}
\caption{Results of \ourmethodshort \hspace{0.1 cm} and other baselines using  , LLaMA-2-13B-Chat on the factoid dataset}
\label{tab:factoid_llama13b}
\end{table*}

\section{Results on CLAPNQ Dataset}
As mentioned in \ref{subsec:clapnq}, here we show the result of \ourmethodshort \hspace{0.1pt} along with Base+RAG and RAFT on the CLAPNQ dataset which is a subset from the Natural Questions dataset.
We see in table \ref{tab:clapnq} that token level recall for RAFT is lower than both the baseline and \ourmethodshort. 

\begin{table*}[]
\center
\resizebox{\textwidth}{!}{
\begin{tabular}{@{}lccc|ccc|ccc|c@{}}
\toprule
\multicolumn{4}{c|}{\textbf{Overall}} &
  \multicolumn{3}{c|}{\textbf{No Overlap}} &
  \multicolumn{3}{c|}{\textbf{Some Overlap}} &
  \multirow{2}{*}{\textbf{Reg. Scores}} \\ \cmidrule(r){1-10}
\multicolumn{1}{l|}{} &
  \textbf{Recall} &
  \textbf{Mixtral-J} &
  \textbf{LLaMA-J} &
  \textbf{Recall} &
  \textbf{Mixtral-J} &
  \textbf{LLaMA-J} &
  \textbf{Recall} &
  \textbf{Mixtral-J} &
  \textbf{LLaMA-J} &
   \\ \midrule
\multicolumn{1}{l|}{\textbf{Base+RAG}} &
  46.7 &
  \textbf{78.0} &
  \textbf{78.0} &
  34.8 &
  65.4 &
  69.2 &
  \textbf{59.6} &
  \textbf{91.7} &
  \textbf{87.5} &
  56.4 \\
\multicolumn{1}{l|}{\textbf{RAFT}} &
  35.7 &
  74.0 &
  \textbf{78.0} &
  25.1 &
  69.2 &
  73.1 &
  47.1 &
  79.2 &
  83.3 &
  -8.4 \\
\multicolumn{1}{l|}{\textbf{PA-RAG}} &
  \textbf{47.7} &
  \textbf{78.0} &
  \textbf{78.0} &
  \textbf{40.4} &
  \textbf{80.8} &
  \textbf{80.8} &
  55.6 &
  75.0 &
  75.0 &
  -1.2 \\ \bottomrule
\end{tabular}
}
\caption{Comparison of \ourmethodshort \hspace{0.1pt} with base model and RAFT on the CLAPNQ dataset to see the effect of trying to ingest already known knowledge.}
\label{tab:clapnq}
\end{table*}

\section{Prompts}
\label{subsec: appendix_prompts}

This section enumerates all the prompts that are used in our experiments.
We use LLaMA-3.3-70B-Instruct\footnote{\url{https://huggingface.co/meta-llama/Llama-3.3-70B-Instruct}} as judge and employ Mixtral-8x22B-Instruct-v0.1\footnote{\url{https://huggingface.co/mistralai/Mixtral-8x22B-Instruct-v0.1}} both for data generation and as a judge to evaluate the fine-tuned model's predicted answers against the ground truth.

\subsection{Prompt for generating question-answer pairs}
The prompt in Table~\ref{tab:qa_prompt} instructs the model to create question-answer pairs from a given document, ensuring that the questions do not use co-referencing or pronouns and that they are fully contextualized.

\begin{table*}[h]
	\centering
     \resizebox{\textwidth}{!}{%
	\begin{tikzpicture}
		\node[draw=black, rounded corners=10pt, fill=gray!20, inner sep=10pt] (box) { 
			\begin{tabular}{p{\linewidth}}
				<s> [INST] Create question answer pairs from the document given below within <document> tags. Title of the document is given in the first line of the document. Do not use co-referencing and pronouns at all in the questions. Do not refer to the document in the question like "according to the document ..." or any similar paraphrasing. When needed, contextualize the question by using the topic that the question is about. You can use the title of the document as well for contextualizing. There are several figures in the document, while referring to the figure in any question, contextualize it by mentioning the title of the passage it was present in. Put questions within <question> and </question> tags and answers within <answer> and </answer> tags. Ensure that the question and answers cover the entire document. When you are done generating QA pairs, generate </done> token. [/INST]\\
			\end{tabular}
		};
	\end{tikzpicture}
 }
	\caption{Prompt for generating question-answer pairs from a document.} 
	\label{tab:qa_prompt}
\end{table*}

\subsection{Prompt for generating multiple answers}
The prompt in Table~\ref{tab:multiple_answers_prompt} enables the generation of multiple distinct answers for a question based on the provided document, encouraging diverse answer styles and formats.

\begin{table*}[h]
	\centering
     \resizebox{\textwidth}{!}{%
	\begin{tikzpicture}
		\node[draw=black, rounded corners=10pt, fill=gray!20, inner sep=10pt] (box) { 
			\begin{tabular}{p{\linewidth}}
				<s> [INST] You are provided with a document within <document> and </document> tags. Followed by the document, you are provided with a question within <question> and </question> tags. Using only the information provided in the document, generate multiple correct, complete and comprehensive answers for the given question, varying the style and format. Each of your answers should be within separate <answer> and </answer> tags so that I can parse and extract them using a code. Each answer must be complete, correct, comprehensive and only from the provided document.  \\
				Start by first generating <answer> token. Then generate one valid answer following any one style or format. Then generate </answer> token. Then again generate <answer> token, followed by another valid answer. Then generate </answer> token. Keep on repeating this till you generate as many correct answers as possible. In the end, generate </done> token once you are done generating all the answers in all possible styles and formats.  \\
				Some example styles and format that you must use:  \\
				1. List Answer: Answer can be in the form of a list.  \\
				2. Extractive answer: Answer with extracted sentences from the document.  \\
				3. Inferential answer: Answer that summarizes the information in the document but is not directly quoted.  \\
				4. Definitions: Answer that gives definitions related to the question.  \\
				5. Examples: Providing examples that can be valid responses to the question.  \\
				Ensure that the answers are not just paraphrases but differ in their content and are factually correct based on the document. [/INST] \\
			\end{tabular}
		};
	\end{tikzpicture}
 }
	\caption{Prompt for generating multiple answers to a question from a document.} 
	\label{tab:multiple_answers_prompt}
\end{table*}

\subsection{Prompt used while fine-tuning} 
Table~\ref{tab:ft_prompt} shows the prompt used while fine-tuning \ourmethodshort\ using RAFT and \ourmethodshort.
Table~\ref{tab:qaft_prompt} shows the prompt for the DSF baseline that trains the model to generate a response to a given question without any retrieved passages in the context. 

\begin{table*}[h]
	\centering
     \resizebox{\textwidth}{!}{%
	\begin{tikzpicture}
		\node[draw=black, rounded corners=10pt, fill=gray!20, inner sep=10pt] (box) { 
			\begin{tabular}{p{\linewidth}}
                    <s> [INST] You are an AI assistant who is provided with a conversation between a user and an agent. User utterances start with "User:" and agent utterances start with "Agent:".  Your task is to generate agent's response to the last user utterance. Enclosed within <passage> tags, you will find various excerpts. These passages may or may not contain the answer. You may or may not use the information in them to generate your response. Present your response within <response> and </response> tokens.
                    
                    <passage\_0>
                    \{document\_0\}
                    </passage\_0>
                    
                    <passage\_1> 
                    \{document\_1\}  
                    </passage\_1>  
                    
                    <passage\_2>
                    \{document\_2\}
                    </passage\_2>
                    
                    <passage\_3>
                    \{document\_3\}
                    </passage\_3>
                    
                    <passage\_4>
                    \{document\_4\}
                    </passage\_4>

                    User: \{data\_identifier\} \{question\}
                    
                    [/INST]
                    
                    \#\#\# Answer:
			\end{tabular}
		};
	\end{tikzpicture}
 }
	\caption{Prompt used while fine-tuning the base model. \{document\_i\} $\forall i = 1\cdots5$ are replaced by the retrieved passages; 
    \{question\} is replaced by the user question. For both book1 and book2, \{data\_identifier\} is replaced by "This query is with reference to IBM Redbooks".
    }
	\label{tab:ft_prompt}
\end{table*}

\begin{table*}[h]
	\centering
     \resizebox{\textwidth}{!}{%
	\begin{tikzpicture}
		\node[draw=black, rounded corners=10pt, fill=gray!20, inner sep=10pt] (box) { 
			\begin{tabular}{p{\linewidth}}
                    <s> [INST] You are an AI assistant who is provided with a conversation between a user and an agent. User utterances start with "User:" and agent utterances start with "Agent:".  Your task is to generate agent's response to the last user utterance. Present your response within <response> and </response> tokens.

                    User: \{question\}
                    
                    [/INST]
                    
                    \#\#\# Answer:
			\end{tabular}
		};
	\end{tikzpicture}
 }
	\caption{Prompt used during Domain-specific fine-tuning (DSF) using only QA pairs without any passages; 
    \{question\} is replaced by the user question..
    }
	\label{tab:qaft_prompt}
\end{table*}

\subsection{Prompt for LLM-as-Judge evaluation}
The prompt in Table~\ref{tab:llm_as_judge_prompt} is used to evaluate the fine-tuned model's predicted answers against the ground truth answers, with Mixtral-8x22B-Instruct-v0.1 and LLaMA-3 70b serving as the evaluation judge.

\begin{table*}[h]
	\centering
    \resizebox{\textwidth}{!}{%
        \begin{tikzpicture}
		\node[draw=black, rounded corners=10pt, fill=gray!20, inner sep=10pt] (box) { 
			\begin{tabular}{p{\linewidth}}
				<s> [INST] You are given a question, the corresponding ground-truth answer and a prediction from a model. Compare the "Ground-truth answer" and the "Prediction" to determine whether the prediction correctly answers the question. \\
                There should be no contradicting statements in a good prediction. The prediction may contain extra information. If the prediction states something as a possibility, treat it as a definitive answer. \\
                A good prediction must contain all the important information presented in the ground truths, but doesn't have to fully match it word by word. \\
                To make your decision, first read the question and Ground-truth answer carefully. Then compare the given Prediction with the Ground-truth answer in the light of the question. \\
                Start with an explanation and reasoning for your evaluation within <explanation> and </explanation> tags. \\
                Then, within <score> and </score> tokens, generate "1" if the Prediction is correct in the light of Ground-truth and question. Otherwise generate "0" if it is incomplete or incorrect.  \\
                Question: \{question\} \\
                Ground-truth answer: \{gold\_response\} \\
                Prediction: \{predicted\_response\} \\
                
                [\slash INST] <\slash s> \\
			\end{tabular}
		};
	\end{tikzpicture}
    };
	\caption{Prompts for Mixtral-8x22B-Instruct-v0.1-based evaluation. Here, \{question\}, \{gold\_response\}, and \{predicted\_response\} are placeholders that are replaced by the actual question, ground truth answer, and predicted answer, respectively.} 
	\label{tab:llm_as_judge_prompt}
\end{table*}

\subsection{Prompt for filtering test dataset}
The prompt in Table~\ref{tab:filtering_prompt} is used to filter dependent questions from the test set of the synthetic datasets. We used Mixtral-8x22B-Instruct-v0.1 to do the filtering.

\begin{table*}[h]
    \centering
    \resizebox{\textwidth}{!}{%
        \begin{tikzpicture}
            \node[draw=black, rounded corners=10pt, fill=gray!20, inner sep=10pt] (box) { 
                \begin{tabular}{p{\linewidth}}
                    <s> [INST] You are given a question, and your task is to determine whether the question is complete and makes sense in isolation. A complete question should not contain text like "based on given example" or "mentioned in the chapter" because only question text is there to answer it, not the chapter or examples.  \\[2pt]
                    
                    Based on this, assign a final score of **Complete** or **Incomplete** for the question. Only assign **Incomplete** if the question uses phrases like these: "based on given example" or "mentioned in the chapter".  \\[2pt]
                    
                    **Scoring:** \\  
                    - **Complete:** The question contains all necessary details. \\  
                    - **Incomplete:** The question uses phrases like "based on given example", "mentioned in the chapter", "mentioned in the passage", "given in the document", or something similar.  \\[2pt]
                    
                    Provide your feedback in the following format: \\  
                    ``` \\ 
                    Feedback:::\\ 
                    Evaluation: (Provide your rationale for the scoring in detail)\\ 
                    Scoring: (State "Complete" or "Incomplete")\\  
                    ```  \\[2pt]
                    
                    **Important:** You must provide values for both 'Evaluation:' and 'Scoring:' in your answer.  \\[2pt]
                    
                    Now, here is the question:  \\  
                    **Question:** {\{question\}}  \\[2pt]
                    
                    Provide your feedback.  \\[2pt]
                    Feedback:::  \\  
                    Evaluation: [/INST] \\
                \end{tabular}
            };
        \end{tikzpicture}
    }
    \caption{Prompt for filtering test data using Mixtral-8x22B-Instruct-v0.1. Here, \{question\} is a placeholder that will be replaced by the actual question.}
    \label{tab:filtering_prompt}
\end{table*}

\section{Example Predictions}

\subsection{Example where the Retriever Fails}

As seen in Table \ref{tab:bad-ret}, \ourmethodshort\ is able to correctly recall the information necessary for answering the question even when the retrieved passages did not contain any relevant information. The RAFT method is expected to perform poorly conditional memorization bias, where as DSF is unable to recall the relevant information.

\begin{table}
	\centering
	\renewcommand{\arraystretch}{1.5}
	\begin{tabular}{l|p{13cm}}
	\toprule
 \textit{Question}&How can extended data tables be retrieved in IBM Storage Virtualize?\\ \midrule
 \textit{Gold Answer}& \colorbox{green}{Many ls commands} in IBM Storage Virtualize can leverage \colorbox{yellow}{the -gui parameter} to retrieve an extended data table similar to the output displayed in the IBM Storage Virtualize GUI. This provides more detailed information for all objects of the requested type and can be helpful for automation and scripting. \\ \midrule
 \textit{Base Model} & IBM Storage Virtualize does not directly support the concept of extended data tables as described in the provided passages, which involve creating external tables in Presto using data in IBM Storage Scale directories. Instead, IBM Storage Virtualize is a software-defined storage solution that provides virtualization of physical storage resources, enabling the consolidation and management of multiple storage systems into a single, unified storage pool.\\
  & To retrieve data from IBM Storage Virtualize, you can use various access methods such as block I/O (iSCSI), file I/O (POSIX compliant), or object storage (S3). The data can then be accessed using appropriate tools or applications based on the chosen access method.\\ 
 & If you need to work with data in a distributed file system like HDFS or use Spark for data processing, you may consider using IBM Watson for Data and AI solutions, such as IBM Watson Studio or IBM Watson Machine Learning, which can interact with IBM Storage Scale and provide the required functionality for managing and processing extended data tables.\\ \midrule
\textit{DSF+RAG} & Extended data tables can be retrieved in IBM Storage Virtualize by using the lsdeltas command with the -extended parameter. This command returns more detailed information about the changed objects, including the old and new values of the object attributes. \\ \midrule
\textit{RAFT} & Extended data tables can be retrieved in IBM Storage Virtualize by using the \colorbox{yellow}{-gui parameter} with the lsdrive command. This will retrieve more detailed information about each drive, including its capacity, status, and type. \\ \midrule
\textbf{\ourmethodshort} & To retrieve extended data tables in IBM Storage Virtualize, you can use the \colorbox{yellow}{-gui parameter} in \colorbox{green}{many ls commands}. This parameter allows you to retrieve an extended data table similar to the output displayed in the IBM Storage Virtualize GUI. For example, the lsdrive command with the -gui parameter will provide more detailed information about each drive, such as UID, block\_size, vendor\_id, product\_id, FRU\_part\_number, FRU\_identity, RPM, firmware\_level, FPGA\_level, quorum\_id, port\_1\_status, port\_2\_status, was\_spare, interface\_speed, write\_endurance\_used, transport\_protocol, compressed, physical\_capacity, physical\_used\_capacity, effective\_used\_capacity, date\_of\_manufacture, protection\_enabled, write\_endurance\_usage\_rate, replacement\_date, anomaly\_detection\_active, and so on. \\
\bottomrule
 \end{tabular}
 \caption{An example where the retriever fails and the models have to recall domain-specific information from their parametric knowledge.}
 \label{tab:bad-ret}
\end{table}

\subsection{Example where the Retriever is Successful}

Refer to Table \ref{tab:good-ret} for responses to a question where the retrieved passages contained the gold answer. We see that although all the methods are able to extract some relevant information from the provided context, they are unable to extract all the necessary information. However, \ourmethodshort\ captures all necessary information from the retrieved passages demonstrating its effectiveness in utilizing the retrieved data for accurate responses.

\begin{table}
	\centering
	\renewcommand{\arraystretch}{1.5}
	\begin{tabular}{l|p{13cm}}
	\toprule
 \textit{Question}&What are the options for configuring password policy in IBM Storage Virtualize?\\ \midrule
 \textit{Gold Answer}& The following attributes can be defined in password policy in IBM Storage Virtualize:
    - \colorbox{green}{Minimum password length}: \colorbox{yellow}{6-64 characters}.
    - Minimum number of:
    - \colorbox{green}{Uppercase characters}: \colorbox{yellow}{0-3}.
    - \colorbox{green}{Lowercase characters}: \colorbox{yellow}{0-3}.
    - \colorbox{green}{Special characters}: \colorbox{yellow}{0-3}.
    - \colorbox{green}{Digits}: \colorbox{yellow}{0-3}.
    - \colorbox{green}{History check} \colorbox{yellow}{(0-10)} before password reuse.
    - \colorbox{green}{Password expiry}: \colorbox{yellow}{0-365 days}.
    - \colorbox{green}{Password expiry warning} \colorbox{yellow}{(0-30 days)} which is displayed on CLI at login only.
    - \colorbox{green}{Password age} \colorbox{yellow}{(1-365 days)} which is the minimum age before a password can change. \\ \midrule
 \textit{Base Model} & The password policy in IBM Storage Virtualize can be configured through the GUI or CLI. Here are the steps to configure it through the GUI:
    1. Select `Settings` > `Security`. 2. Select the `Password Policies` tab. 3. Set the required password parameters in the `Password creation` and `Password expiration and account lockout` sections. 4. Click `Save`. Alternatively, you can use the `chsecurity` command in the CLI to change the password policy. For example, to set the minimum password length to 8 characters, you can use the following command: 
    ```bash IBM\_FlashSystem:ITSO:securityadmin $>$ chsecurity -minpasswordlength 8```. The `chsecurity` command also allows you to configure other password policy attributes such as minimum number of \colorbox{green}{uppercase}, \colorbox{green}{lowercase}, \colorbox{green}{special characters}, \colorbox{green}{digits}, \colorbox{green}{history check}, \colorbox{green}{password expiry}, and \colorbox{green}{password expiry warning}. 
    \dots
    Note that if you disable password reset and you do not have local or remote Security Administrator users except superuser, there is no way to restore management access if superuser password is lost.\\
\midrule
\textit{DSF+RAG} & The options for configuring password policy in IBM Storage Virtualize include setting \colorbox{green}{password expiration}, \colorbox{green}{password expiry warning}, \colorbox{green}{password history check}, \colorbox{green}{minimum password length}, \colorbox{green}{minimum number of uppercase characters}, \colorbox{green}{lowercase characters}, \colorbox{green}{special characters}, \colorbox{green}{digits}, and \colorbox{green}{minimum password age.} \\ \midrule
\textit{RAFT} & The options for configuring password policy in IBM Storage Virtualize include setting \colorbox{green}{minimum password length}, \colorbox{green}{minimum number of uppercase characters}, \colorbox{green}{lowercase characters}, \colorbox{green}{special characters}, and \colorbox{green}{digits, history check before password reuse}, \colorbox{green}{password expiry}, \colorbox{green}{password expiry warning}, and \colorbox{green}{password age}. \\ \midrule
\textbf{\ourmethodshort} & The options for configuring password policy in IBM Storage Virtualize are as follows:

    - \colorbox{green}{Minimum password length}: \colorbox{yellow}{6-64 characters}.
    - Minimum number of:
      - \colorbox{green}{Uppercase characters}: \colorbox{yellow}{0-3}.
      - \colorbox{green}{Lowercase characters}: \colorbox{yellow}{0-3}.
      - \colorbox{green}{Special character}s: \colorbox{yellow}{0-3}.
      - \colorbox{green}{Digits}: \colorbox{yellow}{0-3}.
    - \colorbox{green}{History check} \colorbox{yellow}{(0-10)} before password reuse.
    - \colorbox{green}{Password expiry}: \colorbox{yellow}{0-365} days.
    - \colorbox{green}{Password expiry warning} \colorbox{yellow}{(0-30 days)} which is displayed on CLI at login only.
    - \colorbox{green}{Password age} \colorbox{yellow}{(1-365 days)} which is the minimum age before a password can change.

    These attributes can be set in the GUI by selecting Settings $>$ Security, then selecting the Password Policies tab. The chsecurity command can also be used in the CLI to set the password policy. \\
\bottomrule
 \end{tabular}
 \caption{An example where the retriever is successful and the models must leverage the retrieved information efficiently to answer the questions.}
 \label{tab:good-ret}
\end{table}

\newpage

\section{Licenses}

\begin{table}[h]
\centering
\begin{tabular}{@{}lcl@{}}
\toprule
\multicolumn{1}{c}{\textbf{Tool Used}} & \multicolumn{2}{c}{\textbf{License}}                          \\ \midrule
\textbf{LlamaIndex}                    & \multicolumn{2}{c}{The MIT License Copyright (c) Jerry Liu}   \\
\textbf{LM Evaluation Harness} & \multicolumn{2}{c}{MIT License,  Copyright (c) 2020 EleutherAI} \\
\textbf{HuggingFace}                   & \multicolumn{2}{c}{Apache License, Version 2.0, January 2004} \\ \bottomrule
\end{tabular}%
\caption{Licenses of the different tools we used for running our experiments.}
\label{tab:licenses}
\end{table}

\end{document}